%% file: main.tex
\documentclass[10pt,twocolumn,letterpaper]{article}

\usepackage{cvpr}              
\usepackage[framemethod=TikZ]{mdframed}
\usepackage{graphicx}        
\usepackage{xcolor}          
\usepackage{caption}         
\usepackage{float}           
\usepackage{amssymb}
\usepackage{CJKutf8}
\usepackage{tcolorbox}
\tcbuselibrary{breakable}

\input{preamble}

\definecolor{cvprblue}{rgb}{0.21,0.49,0.74}
\usepackage[pagebackref,breaklinks,colorlinks,allcolors=cvprblue]{hyperref}
\usepackage{algorithm}
\usepackage{algpseudocode}
\usepackage{tcolorbox, lipsum}
\tcbuselibrary{breakable}

\def\paperID{22449} 
\def\confName{CVPR}
\def\confYear{2025}

\title{Do Audio-Visual Large Language Models Really See and Hear?}

\author{
Ramaneswaran Selvakumar\raisebox{0.5ex}{\small *}\quad Kaousheik Jayakumar\raisebox{0.5ex}{\small *}\quad S Sakshi\\
Sreyan Ghosh\quad Ruohan Gao\raisebox{0.5ex}{\small \#}\quad Dinesh Manocha\raisebox{0.5ex}{\small \#}\\
\small \raisebox{0.5ex}{\small *}Equal contribution \quad \raisebox{0.5ex}{\small \#}Equal advising\\
University of Maryland, College Park \\
\small Project webpage: \href{https://ramaneswaran.github.io/avllm_interpretability/}{\texttt{ramaneswaran.github.io/avllm\_interpretability}}}

\begin{document}

\maketitle

\input{sec/0_abstract}    
\input{sec/1_intro}
\input{sec/2_related_work}

\input{sec/3_preliminary}
\input{sec/4_investigating_attention}

\input{sec/5_investigating_audio_representations}

\input{sec/6_investigating_information_flow}

{
    \small
    \bibliographystyle{ieeenat_fullname}
    \bibliography{main}
}

\input{sec/X_suppl}

\end{document}

%% file: preamble.tex
\definecolor{nicegreen}{rgb}{0.1, 0.6, 0.2}

%% file: sec/0_abstract.tex
\begin{abstract}

Audio-Visual Large Language Models (AVLLMs) are emerging as unified interfaces to multimodal perception. We present the first mechanistic interpretability study of AVLLMs, analyzing how audio and visual features evolve and fuse through different layers of an AVLLM to produce the final text outputs. We find that although AVLLMs encode rich audio semantics at intermediate layers, these capabilities largely fail to surface in the final text generation when audio conflicts with vision. Probing analyses show that useful latent audio information is present, but deeper fusion layers disproportionately privilege visual representations that tend to suppress audio cues. We further trace this imbalance to training: the AVLLM’s audio behavior strongly matches its vision-language base model, indicating limited additional alignment to audio supervision. Our findings reveal a fundamental modality bias in AVLLMs and provide new mechanistic insights into how multimodal LLMs integrate audio and vision.
\end{abstract}

%% file: sec/1_intro.tex
\section{Introduction}
\label{sec:intro}

Audio-Visual Large Language Models (AVLLMs)~\cite{xu2025qwen2, chen2025omnixr, sun2024video, zhang2023video, lyu2023macaw} extend large language models (LLMs) to process real-world multimodal inputs such as audio and video. Audio conveys information that vision alone cannot: off-screen events, speech, music, and ambient cues are primarily auditory. By combining these signals with the reasoning capabilities of LLMs, AVLLMs achieve a more holistic understanding of complex scenes--enabling applications in multimedia analysis, education~\cite{AlShaikh2024}, human–computer interaction~\cite{chen2025emova, zhou2025daily}, robotics~\cite{hirose2025omnivla, yao2025roboego}, healthcare~\cite{mesko2023impact}, and biodiversity monitoring~\cite{muller2023soundscapes}.

\begin{figure}
     \centering
     \includegraphics[width=0.95\linewidth]{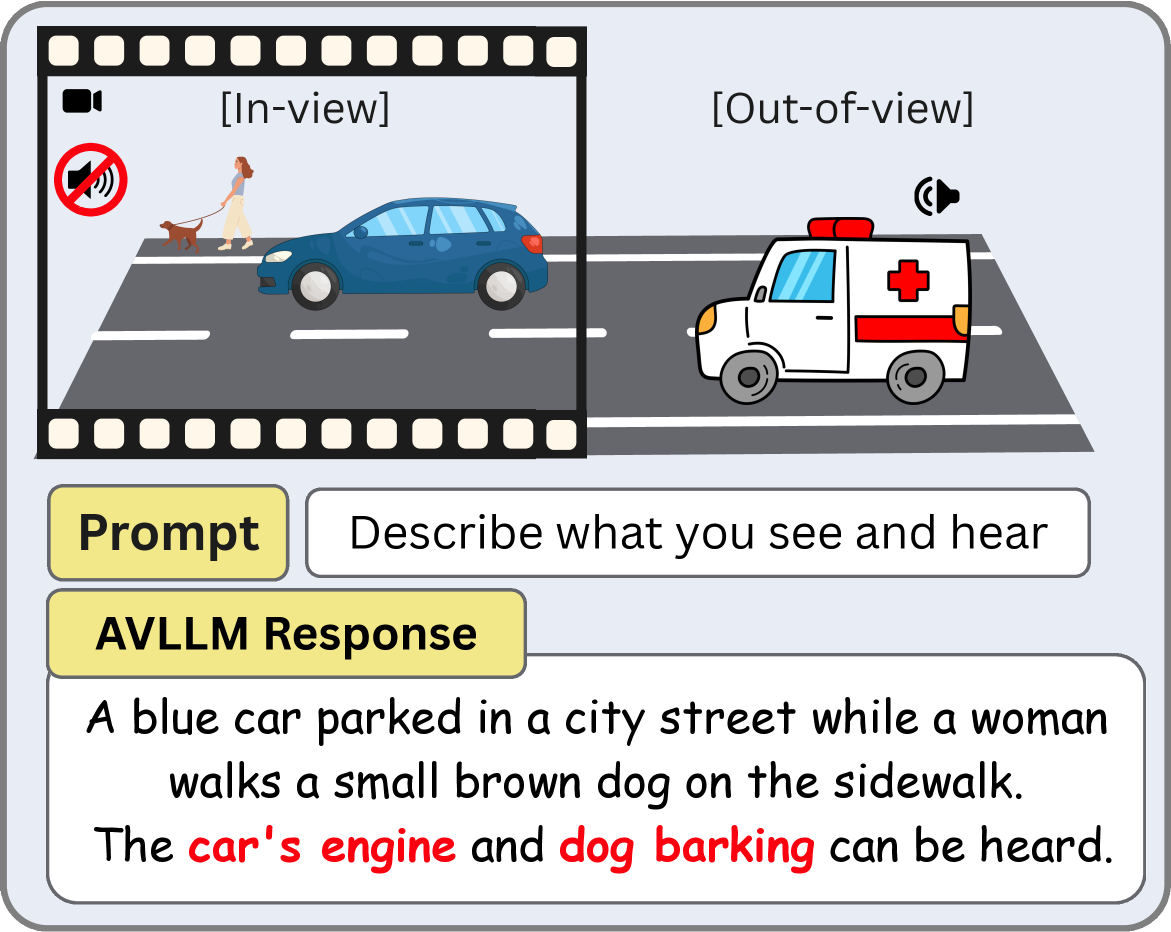}
     \caption{\textbf{Illustration of visual bias.} AVLLMs exhibit a critical modality bias, often prioritizing visual cues over vital audio cues. The diagram illustrates a counterfactual scene, visible objects (a blue car and a woman walking a dog) are silent and the only audible sound is an out-of-view ambulance siren. When prompted to describe the scene, the AVLLM hallucinates audio events (car engine, dog barking) and misses the actual siren sound.
     }
\label{fig:teaser}
 \end{figure}

\begin{figure}
     \centering
     \includegraphics[width=1\linewidth]{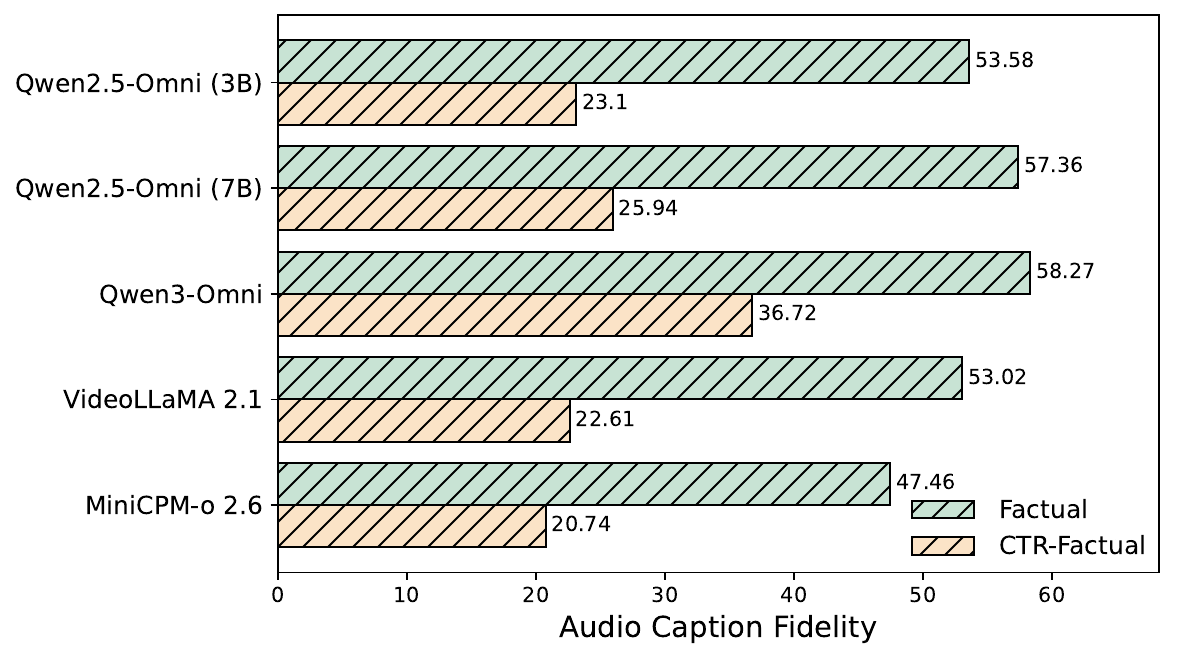} \vspace{-0.4cm}
     \caption{\textbf{Audio Understanding Performance.} Audio understanding severely degrades under audio-visual conflict.
     }
\label{fig:audio_understanding_performance}
 \end{figure}

Despite rapid progress, current AVLLMs remain black-boxes in terms of how they process and utilize audio-visual information. While interpretability has been explored for text-only LLMs~\cite{transcoder_llm_interp_nips, gurnee2023finding, zhang2023towards}, vision-language~\cite{kaduri2025s, neotowards, ben2024lvlm}, and audio-language models~\cite{yang2025audiolens, paek2025learning, ma2025behind}, the mechanisms of audio-visual integration have not been studied. This opacity raises reliability concerns, particularly in safety-critical settings---\eg, an autonomous vehicle should respond to an out-of-view ambulance siren even when it is not visible. To illustrate this, consider Fig~\ref{fig:teaser}: a scene where a blue car and a woman walking a dog are visible, while an ambulance siren is heard off-screen. When prompted to ``describe what you see and hear,'' existing AVLLMs frequently hallucinate sounds from visible but silent objects and ignore the siren. We observe that for current AVLLMs (Fig~\ref{fig:audio_understanding_performance}) relative performance drops by up to 56\% on such counter-factual samples with conflicting modalities, as compared to factual samples where both modalities are aligned, indicating that current AVLLMs rely heavily on vision and underutilize audio cues.

To address these issues, we systematically analyze why AVLLMs fail to utilize audio inputs effectively. Most AVLLMs adopt an adapter-based architecture~\cite{caffagni-etal-2024-revolution}, where pretrained audio and vision encoders feed representations through learned adapters into the LLM token space-extending designs such as Llava~\cite{liu2023visual}. We focus on the LLM backbone, the largest and most influential component, to examine how audio and visual representations evolve, interact, and influence text generation across layers. Unlike prior interpretability work on single-modality models~\cite{kaduri2025s, neotowards, ben2024lvlm, yang2025audiolens}, AVLLMs present unique challenges: audio and vision interact through complex cross-modal pathways, and their complementarity makes isolating each modality’s contribution far more difficult.

Using this testbed, we perform a multi-stage mechanistic interpretability analysis to understand where and why audio information fails to manifest in generation: \textbf{(1)} We analyze attention patterns across layers to identify where AVLLMs allocate focus between modalities and whether audio receives sufficient attention. \textbf{(2)} We then probe audio representations to determine if they encode meaningful information. \textbf{(3)} To establish causal relationships between modality representations and outputs, we design attention knockout experiments that selectively block audio or visual pathways, allowing us to trace how each modality influences generation. \textbf{(4)} Finally, to investigate the origin of any observed bias, we compare output token distributions with base vision-language models, testing whether visual dominance stems from inherited training artifacts or architectural constraints. Together, these analyses provide the first mechanistic evidence of how and where modality imbalance arises in AVLLMs. To summarize, our main contributions are:

\begin{enumerate}
    \item We present the first systematic mechanistic analysis of Audio-Visual Large Language Models, dissecting how audio and visual representations are encoded, integrated, and manifested in text generation.
    \item We show that AVLLMs’ audio understanding drastically deteriorates by upto 56\% when audio and visual cues conflict--revealing a strong modality preference despite architectural capacity for multimodal reasoning.
    \item By probing audio token representations in intermediate layers we demonstrate that they evolve into interpretable onomatopoeic tokens which describe sound events, indicating that AVLLMs encode strong latent audio semantics that remain untapped at generation time.
    \item Using causal mediation via attention knockouts, we find that deeper layers prioritize visual features and actively suppress audio information. Blocking visual signals at these layers restores audio reasoning, providing direct causal evidence of cross-modal interference.
    \item We show that the output token distributions of AVLLMs strongly mirror those of their base vision-language models, suggesting that the visual bias potentially originates from its alignment tuning and data rather than architectural limitations.
\end{enumerate}

%% file: sec/2_related_work.tex
\section{Related Work}

\paragraph{Audio Visual Large Language Models (AVLLMs).} AVLLMs aim to integrate auditory and visual perception in large language models, thus enabling reasoning across audio, vision, and text. 
Most AVLLMs follow an adapter-based architecture, popularized by Large Vision Language Models (LVLMs)~\cite{caffagni-etal-2024-revolution, bordes2024introductionvisionlanguagemodeling} where frozen audio and visual encoders are connected to a pretrained language backbone through learned projection modules (adapters). Early systems such as Video-LLaMA \cite{zhang2023video} and PandaGPT \cite{su2023pandagpt} adopt simple MLP adapters to map modality embeddings into the LLM token embedding space. More recent AVLLMs such as Qwen2.5-Omni \cite{xu2025qwen2}, Qwen3-Omni \cite{yang2025qwen3}, and MiniCPM \cite{hu2024minicpm} have dedicated modules that temporally align audio and visual features before mapping to the LLM embedding space, improving synchronization and contextual grounding. While most approaches build on top of base LLMs, others, such as InternOmni~\cite{internomni2024blog} build on existing LVLMs like InternVL~\cite{chen2024internvl}. This extension strategy improves training efficiency and leverages strong vision-language foundations, but may introduce visual bias from the underlying LVLM.

\vspace{-0.1in}
\paragraph{Mechanistic Interpretability of LLMs.} Mechanistic interpretability seeks to understand the internal mechanisms within neural networks. For LLMs, this has revealed how models store factual knowledge~\cite{nanda2023fact, meng2022locating, geva2021transformer}, localize specific capabilities~\cite{wanginterpretability, olsson2022context}, and process information across layers~\cite{elhage2021mathematical}. While well-established for text-only models, mechanistic interpretability remains nascent for multimodal LLMs. \cite{neotowards} show that object information is localized in token representations and gradually aligns with language representations, while \cite{kaduri2025s} and \cite{zhang2025cross} trace cross-modal image token flow, revealing a distinct two-step processing pattern.

\vspace{-0.15in}
\paragraph{Tools For Mechanistic Interpretability.} Understanding and explaining neural networks, particularly language models is crucial for identifying their behavior and limitations. A widely used approach is causal mediation analysis~\cite{vig2020causal, mueller2024quest}, which attributes the contribution of key components, often employing knock-out techniques to assess the impact of removing specific elements. Another approach is to probe the internal representations using linear classifiers~\cite{kumar2022probing, alain2016understanding}, an extension of which is logit lens~\cite{nostalgebraist2020interpreting} where the the language model's unembedding layer is used to probe representations. In our work, we use the logit lens to probe whether semantic audio information is preserved in intermediate representations and further utilize attention knockout, a form of causal mediation analysis to trace the cross-modal flow of audio-visual information in the LLM.

%% file: sec/3_preliminary.tex
\section{Preliminary}

\noindent\textbf{Transformer Language Models.} Transformer-based language models consist of: a tokenizer, an embedding layer, a transformer backbone comprising stacked transformer layers and an unembedding layer. An auto-regressive transformer language model takes as input a tokenized sequence $X = (x_1, ..., x_n)$ and outputs a probability distribution over vocabulary $V$ to predict the next token $x_{n+1}$. 

The embedding layer maps each input token $x_i$ to a corresponding vector $e_i^t \in \mathbb{R}^{d_m}$ forming the sequences of text embeddings $E_t = (e_1^t, ..., e_n^t)$. These embeddings are then refined through a series of transformer layers, each consisting of Multi-head Self-Attention (MHA) and Feed-forward Network (FFN) sublayers with residual connections:
\begin{align}
h'^l_i &= h^{l-1}_i + a^l_i \\
h^l_i &= h'^l_i + f^l_i
\end{align}

\noindent where $h^l_i$ is the representation of token $x_i$ at layer $l$, $a^l_i$ is the output from the MHA sublayer, and $f^l_i$ is the output from the FFN sublayer. All vectors $h^l_i, a^l_i, f^l_i \in \mathbb{R}^{d_m}$.

The MHA sublayer enables information flow between tokens by computing weighted aggregations based on learned query-key similarities. Each token selectively attends to tokens from other positions in the sequence, allowing the model to build contextualized representations:
\begin{equation}
\text{MHA}(Q, K, V) = \text{softmax}\left(\frac{QK^T}{\sqrt{d_k}} + M\right)V.
\end{equation}
Here $Q$, $K$, $V$ are query, key and value matrices derived through learned projections and $M$ is a mask applied to enforce the causal nature of the auto-regressive model, where each position can only attend to previous positions and itself. The mask is defined as: 
\begin{equation}
M_{ij} = \begin{cases}
0 & \text{if } i \geq j \\
-\infty & \text{otherwise}
\end{cases}
\end{equation}
After passing through $L$ transformer layers, the final representation $h_i^L$ of each token is passed through an unembedding layer (language modeling head) to produce logits over the vocabulary $z_i$ given by: 
\begin{equation}
z_i = W_U h^L_i + b,
\end{equation}
where $W_U \in \mathbb{R}^{|V| \times d_m}$ is the unembedding matrix and $b$ is a bias term. A softmax operation converts these logits into a probability distribution over the vocabulary to predict the next token. 

\vspace*{0.5em}
\noindent\textbf{Audio-Visual LLMs} extend transformer language models to process audio and visual inputs alongside text. They consist of: (i) a pre-trained vision encoder that processes images by dividing them into patches, with each patch embedded into a vector (ii) a pre-trained audio encoder that similarly processes audio spectrograms by dividing them into patches and embedding each patch, and (iii) learned adapter modules that project these modality-specific features into the LLM's representation space. Formally, visual features are projected to obtain visual embeddings $E_V = (e_1^V, ..., e_m^V)$ and audio features are projected to obtain audio embeddings $E_A = (e_1^A, ..., e_k^A)$ where $e_i^V, e_i^A\in \mathbb{R}^{d_m}$. These are concatenated with $E_t$ to form the complete input sequence:
\begin{equation}
E = [E_V, E_A, E_t]
\end{equation}
In practice, audio and visual embeddings are often interleaved within the sequence. This architecture allows AVLLMs to process multimodal information within a unified framework, leveraging the powerful reasoning capabilities of large language models to integrate and reason over audio-visual inputs.

\section{Experimental Setup}

\textbf{Evaluation Task.} We evaluate AVLLMs using the audio-visual captioning task with an input prompt: ``describe what you see and hear''. This fundamental task directly probes a model's ability to perceive and integrate information from both modalities. Alternative tasks such as multi-choice or binary question answering offer limited insight into underlying reasoning processes, and their outputs can often be explained by superficial pattern matching rather than genuine reasoning~\cite{balepur2024artifacts, Balepur2024IsYLA, Huang2024MMEvalProCMA}. In contrast, natural language captions are directly interpretable: outputs are human-readable, allowing us to assess what information each modality contributes and identify specific failure modes. Later in our analysis we further probe modality-specific capabilities using targeted instructions such as ``describe what you see'' and ``describe what you hear'' to isolate visual and audio reasoning independently.

\noindent\textbf{Evaluation Dataset.} In natural videos, audio and visual content are highly correlated, making it difficult to determine whether models genuinely process each modality independently or rely predominantly on one to infer the other. Ideally, we would test scenarios where all audible sounds originate from out-of-view sources. While such cases occur naturally, they are rare and difficult to scale for systematic evaluation. Following prior work~\cite{sung2024avhbench, chowdhury2025avtrustbench} we instead construct counterfactuals by deliberately mismatching audio and visual content by swapping original audio track with an alternative. Such counterfactual samples have been used successfully to stress-test LLMs~\cite{Lee2024VLindBenchMLA, Nguyen2024SFRRAGTCA, Liu2023RECALLABA}. We source our samples from AudioCaps~\cite{audiocaps}, a standard benchmark for audio captioning where videos are derived from YouTube and audio captions annotated by humans. In the end, we curate an evaluation set of 500 samples, with equal number of factual and counter-factual videos. We note that existing audio-visual benchmarks are insufficient for our analysis. For instance, benchmarks like DailyOmni \cite{zhou2025daily} couple perception with reasoning, making it difficult to isolate perceptual capabilities. Other perception based benchmarks like AVHBench \cite{sung2024avhbench} does not require genuine cross-modal reasoning, as outputs can often be derived from a single modality. (Refer to Supplementary A for more details).

\vspace{-2pt}
\begin{figure}
     \centering
     \includegraphics[width=1\linewidth]{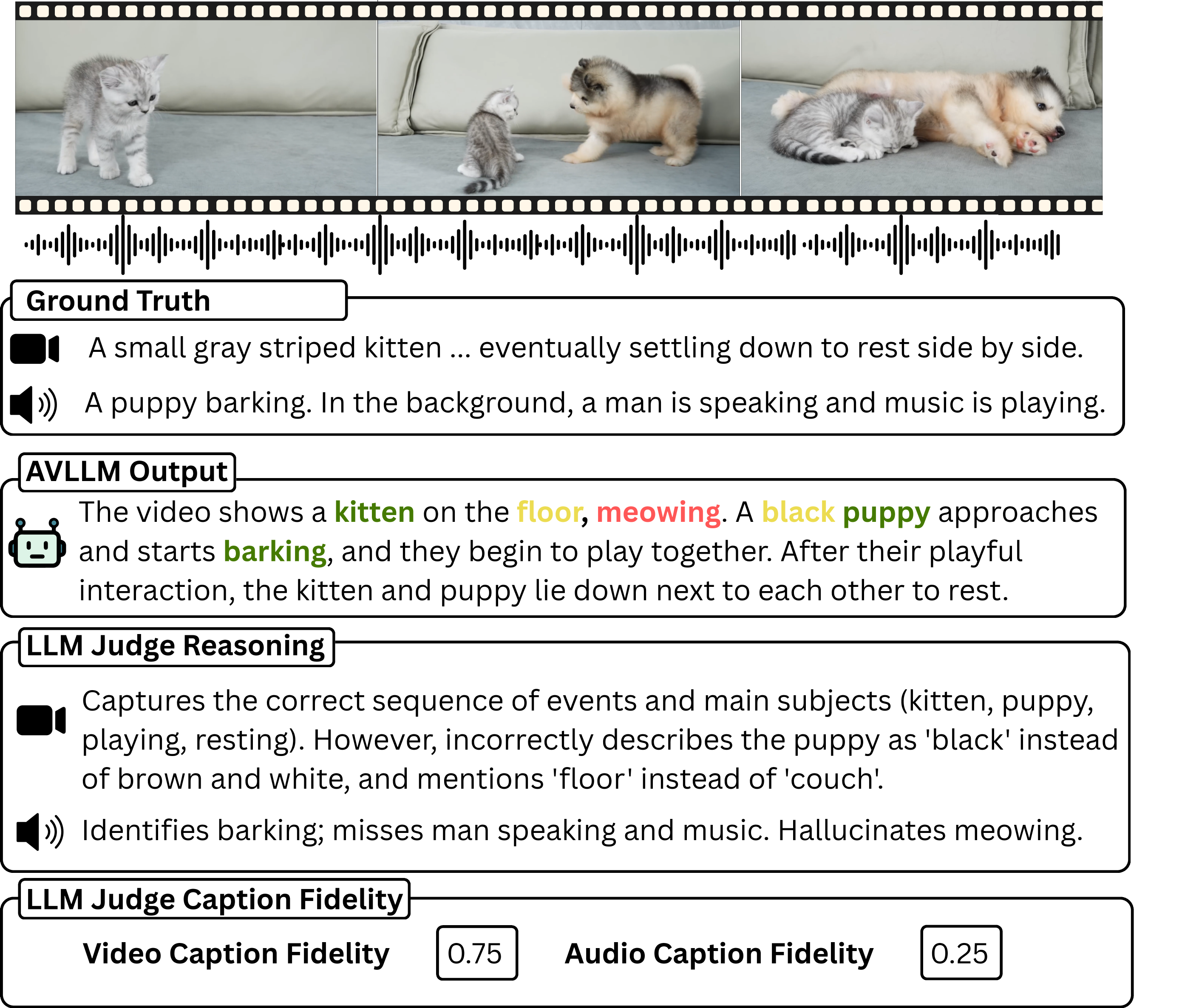} \vspace{-0.4cm}
     \caption{\small \textbf{Caption evaluation.} We use an open-source reasoning LLM to evaluate audio-visual captions by assessing temporal sequences, object attributes, and cross-modal relationships. This approach is interpretable (explicit reasoning for scores) and flexible, we can calibrate it using in-context examples.
     }
\label{fig:caption_fidelity_metric}
 \end{figure}
\vspace{-2pt}
\begin{figure}
     \centering
     \includegraphics[width=1\linewidth]{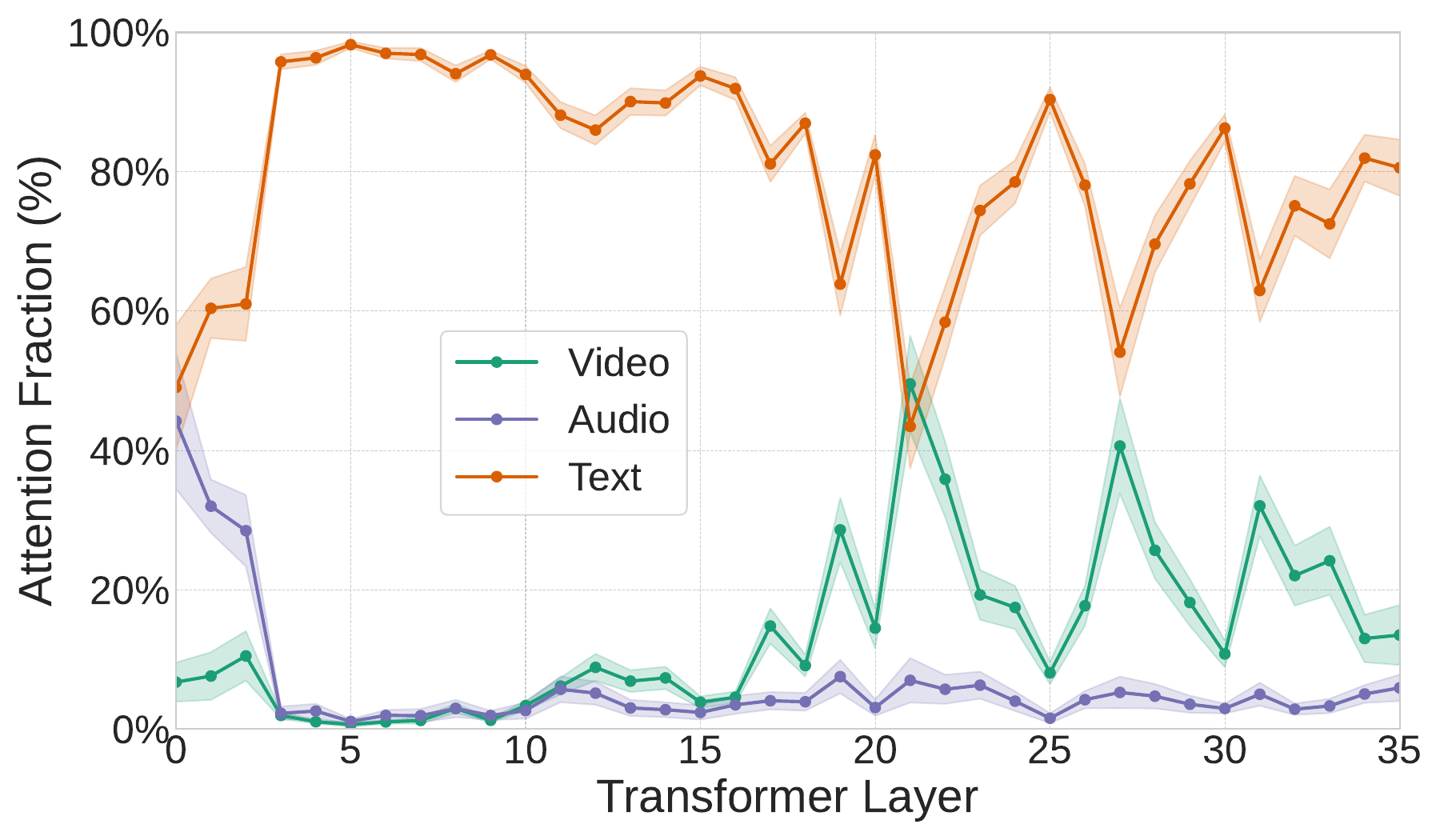} \vspace{-0.7cm}
     \caption{\textbf{Mean attention from generated to input tokens.} Generated tokens allocate high attention to audio in early layers (40-50\% in layers 0-5), which drops to near-zero afterward. Video attention steadily increases through deeper layers, reaching 20-40\% in layers 15-30.
     }
\label{fig:attention_across_layers}
 \end{figure}

\noindent\textbf{Evaluation Models.} We primarily use Qwen2.5-Omni 3B~\cite{xu2025qwen2} in our main experiments. We further validate the generalizability of our findings on additional models such as Qwen3-Omni~\cite{xu2025qwen3omnitechnicalreport}, Video-LLaMA~\cite{zhang2023video}, and MiniCPM-o~\cite{hu2024minicpm} (Refer to Supplementary D for additional results).

\vspace*{0.5em}
\noindent\textbf{Audio Visual Caption Evaluation.} Evaluating generated captions requires measuring fidelity to ground-truth audio and visual captions. Traditional metrics (BLEU~\cite{papineni2002bleu}, METEOR~\cite{banerjee2005meteor}, CIDEr~\cite{banerjee2005meteor}, ROUGE~\cite{lin2004rouge}) fail to capture semantic variability, while embedding-based approaches like CLIP-score~\cite{hessel2021clipscore} behave as bag-of-words models~\cite{yuksekgonuland}, missing compositional structure~\cite{kamath2023text}. On the other hand, methods for evaluating static image captions~\cite{rohrbach2018object, kaduri2025s} do not transfer to videos, which involve temporal dynamics, state changes, and audio-visual correspondence that cannot be evaluated using highly-engineered approaches. 

Prior work shows that LLMs can serve as effective judges for evaluating free-form text when given clear rubrics and references~\cite{chan2023clair}. Building on this, we employ an LLM judge to evaluate caption fidelity by reasoning over ground-truth audio and visual captions (Fig.~\ref{fig:caption_fidelity_metric}). We prompt the model to assess generated descriptions by giving a score between 0 and 1 for each modality separately. Before scoring, the judge first performs deep reasoning over specific details such objects, actions, temporal ordering, and audio events. We calibrate ratings using few-shot in-context examples and request summarized reasoning alongside scores, resulting in interpretable fidelity measurements. We implement this LLM-as-judge using Qwen3-32B~\cite{yang2025qwen3}. Human study on this metric shows strong correlation with human judgements: Spearman $\rho = 0.816$ for audio caption fidelity and Spearman $\rho = 0.732$ for video caption fidelity. (Refer to Supplementary B for more details)

%% file: sec/4_investigating_attention.tex
\section{Investigating Attention Pattern}

We observed severe degradation in audio understanding performance for counterfactual samples, suggesting that in naturally aligned videos, performance may be largely driven by vision. This raises a fundamental question: do AVLLMs attend to audio inputs at all, or do they effectively ignore audio representations during generation? To investigate this, we analyze attention patterns across transformer layers during audio-visual captioning. We prompt models to "describe what you see and hear" and track the average attention allocated by generated tokens to three input token types: video tokens, audio tokens, and query text tokens. We compute these patterns across all layers over our evaluation set and visualize the results in Figure~\ref{fig:attention_across_layers}.

We observe three key patterns. First, query tokens dominate attention across all layers, capturing 60-100\% despite comprising the fewest tokens. Second, audio tokens receive surprisingly high attention (40-50\%) in early layers (0-5), but this attention drops to near-zero in subsequent layers. Third, video tokens show the opposite pattern, their attention steadily increases through middle layers (15-30), reaching 20-40\%. This creates a striking asymmetry in deeper layers, where vision receives substantial attention while audio is largely ignored.

\begin{mdframed}[linewidth=1pt, linecolor=black, leftmargin=1pt, rightmargin=1pt, innerleftmargin=10pt, innerrightmargin=10pt, innertopmargin=4pt, innerbottommargin=2pt, backgroundcolor=gray!20, roundcorner=5pt]
\textit{\textbf{Key Takeaways:}} \small The high audio attention in early layers confirms that AVLLMs do attend to audio representations. However, this attention is concentrated in early layers (0-5) and drops to near-zero afterward. In contrast, vision receives increasing attention in deeper layers (15-30). This reveals an asymmetry: audio is primarily attended to in the early layers, while vision dominates attention in the later layers. 
\end{mdframed}

%% file: sec/5_investigating_audio_representations.tex
\section{Probing Audio Representations}
\vspace{-2pt}
\begin{figure}
     \centering
     \includegraphics[width=1\linewidth]{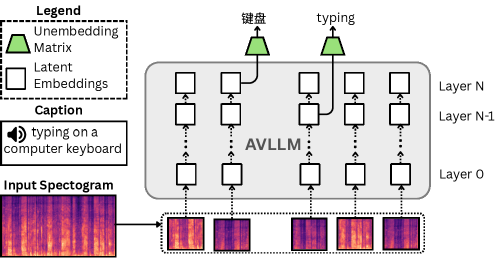} \vspace{-0.7cm}
     \caption{\textbf{Probing Audio Representations.} We decode intermediate layer audio representations using the base LLM's unembedding matrix and observe that they decode into meaningful concepts describing sound events and their sources, in multiple languages (e.g., \begin{CJK}{UTF8}{gbsn}键盘\end{CJK}/keyboard, typing).
     }
\label{fig:logit_lens}
 \end{figure}
\vspace{-2pt}
\begin{figure*}[t]
    \centering
    \includegraphics[width=\textwidth]{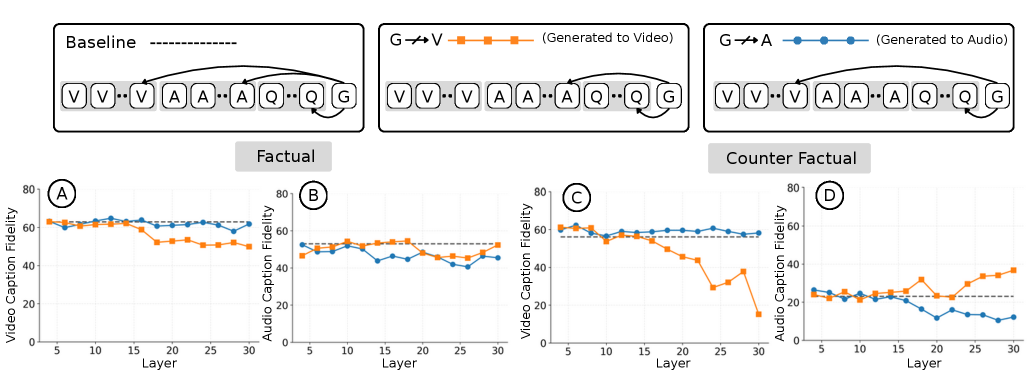}
    \vspace*{-2em}
    \caption{\textbf{Blocking audio-visual information flow to generated tokens} We block attention from generated tokens to vision (\textcolor{orange}{G$\nrightarrow$V}) or audio (\textcolor{blue}{G$\nrightarrow$A}) and measure impact on caption fidelity. 1) \textcolor{blue}{G$\nrightarrow$A} for factual samples does not degrade video understanding\textcircled{A}  as expected (ideally audio cues should not be used for video understanding) 2) surprisingly audio understanding\textcircled{B} does not degrade either, (AVLLM instead correlates audio cues from vision). On the contrary, \textcolor{orange}{G$\nrightarrow$V}, especially in deeper layers, in counterfactual samples, improves audio understanding \textcircled{D} (misleading vision features dominate over audio in cross-modal integration to generated text)} 
    \label{fig:knockout_generated_single_modality}
\end{figure*}

While attention analysis confirms that AVLLMs attend to audio in early layers, it does not reveal whether these representations contain meaningful audio information. To investigate 
this, we directly probe audio representations using logit lens (Fig.~\ref{fig:logit_lens}). This technique decodes hidden states at each audio token position using the model's unembedding matrix $W_U$, 
projecting them into probability distributions over the vocabulary. For each layer $l$, we extract the token with the highest logit at each audio token position $i$ from the hidden state $h_i^l$.

We observe that these decoded tokens meaningfully capture sound sources and event attributes present in the audio. These representations evolve primarily in the later layers, with meaningful audio information emerging in the last 5 layers. We observe that sound sources (e.g., 'drill', 'engine', 'horse', 'keyboard') are more consistently represented, while actions (e.g., 'drilling', 'typing', 'neighing') appear less consistently and sound attributes (e.g., 'light music', 'rock music') are rare.

Interestingly, these sound related tokens often appear in multiple languages, such as \begin{CJK}{UTF8}{gbsn}马\end{CJK} (horse), \begin{CJK}{UTF8}{gbsn}键盘\end{CJK} (keyboard), and \begin{CJK}{UTF8}{gbsn}门 \end{CJK}(door). Consistent with \cite{neotowards}, we note that this is surprising as prior work~\cite{hinck-etal-2024-llava, wendler-etal-2024-llamas} indicates that LVLMs pre-dominantly use English as the latent language in their intermediate representations. However, in our experiment, the base model (Qwen) is multilingual and pre-trained on substantial Chinese data, and as a consequence, the learned audio representations map to concepts in multiple languages.

These decoded tokens enable us to measure latent audio understanding. We extend our LLM-as-judge framework to evaluate whether decoded tokens capture reference audio events, providing a measure of recall using internal representations. Earlier, we observed Qwen2.5Omni achieves only 23\% audio caption fidelity on counterfactual samples (compared to 60\% on factual samples), suggesting models fail to understand audio when visual cues conflict. However, we observe that  Qwen2.5Omni achieves a latent audio understanding score of 61.4\%. Moreover, we observe that even in cases where generated text completely omits the correct audio events, these events are present in internal representations. This gap between latent and manifested audio understanding suggests that the poor understanding does not stem from lack of meaningful representations rather a failure to incorporate them during the generation process.

\begin{tcolorbox}[breakable]
\textit{\textbf{Key Takeaways:}} \small Probing representations of audio token reveals that AVLLMs encode meaningful audio semantics internally (61.4\% latent understanding), yet generated captions achieve only 23\% audio fidelity for counterfactual samples. This gap reveals that poor audio performance doesn't stem from lack of meaningful representations, but from a failure to integrate these during the generation process.
\end{tcolorbox}

%% file: sec/6_investigating_information_flow.tex
\section{Investigating Information Flow}

Our probing experiments reveal that meaningful audio representations are present in AVLLMs, but they do not consistently manifest in final generated text. To understand where audio information fails to propagate, we trace its flow through the network layers. In transformer-based models, attention mechanisms serve as the primary pathways for information flow~\cite{elhage2021mathematical}. In AVLLMs, these pathways enable cross-modal interactions between audio, visual, and text representations. By analyzing these attention pathways, we can identify where audio information integrates into the output and where this integration fails.

To trace information flow, we conduct causal mediation analysis using attention knockout. We block attention from source tokens to target tokens at different layers. If blocking this attention causes the output text to change, this indicates information from the source was influencing the generation. We apply this method to both audio and video tokens as sources, targeting text generation tokens. This allows us to identify where each modality integrates into the final output and potentially reveals whether one modality interferes with the other during cross-modal integration.

Formally, we implement attention knockout by modifying the attention mask to prevent queries at target positions from attending to keys and values at source positions. Let $S$ denote the set of source token positions (tokens being blocked from
being attended to) and $T$ denote the set of target token positions (tokens whose attention is being blocked). To prevent information flow from $S$ to $T$ at layer $l$ and attention head $j$, we modify the attention mask by setting $M_{s,t}^{(l,j)} = -\infty$ for all $s \in S$ and $t \in T$, which causes the
softmax attention weights to become approximately zero:

\begin{equation}
M^{(l, j)}_{s, t} =
\begin{cases}
-\infty &\text{if } s \in S \text{ and } t \in T \\
0 &\text{otherwise}
\end{cases}
\end{equation}

We denote this intervention as $T \nrightarrow S$, indicating that tokens at positions $T$ cannot attend to tokens at positions $S$. For example, $G \nrightarrow A$ represents blocking generated text tokens from attending to audio tokens, where $S$ and $T$ contain audio and generated text token positions respectively.

\noindent\textbf{Experimental Setup} We block attention from generated text tokens to either audio or video tokens at different layers. For each layer $l$, we apply the knockout over a window of 9 consecutive layers centered around $l$ and measure the impact on the generated caption. To better isolate and study individual modalities, we use modality-specific prompts to measure video understanding (``describe what you see'') and audio understanding (``describe what you hear''). We evaluate on both factual (aligned audio-visual) and counterfactual (mismatched audio-visual) samples.

\subsection{Factual Audio-Visual Understanding}

\noindent\textbf{Video Understanding} Figure.~\ref{fig:knockout_generated_single_modality}A shows attention knockout results for video understanding in the factual setting. As expected, blocking audio (G$\nrightarrow$A) produces no significant degradation as the model appropriately relies on visual information for visual tasks. Blocking video (G$\nrightarrow$V), we observe a moderate drop of $\sim10\%$ starting from middle layers (L15). However, this degradation is largely mitigated as the model compensates by leveraging audio information to describe the scenes, demonstrating audio-visual complementarity in action.

\noindent\textbf{Audio Understanding} Figure.~\ref{fig:knockout_generated_single_modality}B reveals more intriguing patterns. When we block audio pathways (G$\nrightarrow$A), performance drops by only $\sim10\%$ which is far less than expected for an audio-specific task. This indicates the model infers audio content from visual cues rather than directly using audio cues. More surprisingly, G$\nrightarrow$V also produces a $\sim10\%$ drop, contrasting with the video understanding case where blocking audio had no effect. This asymmetry suggests that vision influences audio processing even in naturally aligned settings, hinting at systematic visual dominance.

Moreover, these results reveal strong audio-visual complementarity: each modality compensates when the other is blocked, making it difficult to isolate genuine modality-specific processing. This further motivates our counterfactual evaluation, where mismatched audio-visual content forces the model to rely on each modality independently.

\subsection{Counter-Factual Audio-Visual Understanding}

\noindent\textbf{Video Understanding} Figure~\ref{fig:knockout_generated_single_modality}C shows counterfactual video understanding results. In this setting, mismatched audio-visual content eliminates complementary information between modalities. As expected, blocking audio (G$\nrightarrow$A) produces no performance degradation, and audio does not mislead or interfere with visual information processing even when semantically unrelated. Blocking video (G$\nrightarrow$V), however, reveals clear evidence of video integration patterns. Performance degrades gradually starting in middle layers (L15), with severe drops ($\sim40\%$) in the final layers. Without audio compensation available, these results confirm that cross-modal video information transfer begins in middle layers and concentrates heavily in deeper layers.

\vspace*{0.5em}
\noindent\textbf{Audio Understanding} Figure~\ref{fig:knockout_generated_single_modality}D exposes the critical asymmetry in modality processing. Upon blocking audio (G$\nrightarrow$A), audio understanding degrades substantially-gradually in middle layers and severely in final layers (losing up to $\sim50\%$ relative performance). This confirms that audio information also transfers primarily in the deepest layers of the network, mirroring the integration pattern observed for video. However, blocking video( G$\nrightarrow$V) produces a striking result: audio understanding actually improves, recovering approximately $50\%$ relative performance in the final layers and approaching factual-setting levels. This demonstrates that vision actively interferes with audio processing in these deep integration layers and blocking visual pathways recovers the model's latent audio understanding capabilities. We observe similar results on other models too which can be found in Supplementary D. 

\begin{tcolorbox}[breakable]
\textit{\textbf{Key Takeaways:}} \small Both audio and video integrate in deeper layers of the network. However, vision actively dominates audio in the final integration stage. Blocking visual pathways in these layers recovers the model's latent audio understanding capabilities.
\end{tcolorbox}

\section{Investigating Origins of Visual Bias}

\begin{figure*}[t]
    \centering
    \includegraphics[width=\textwidth]{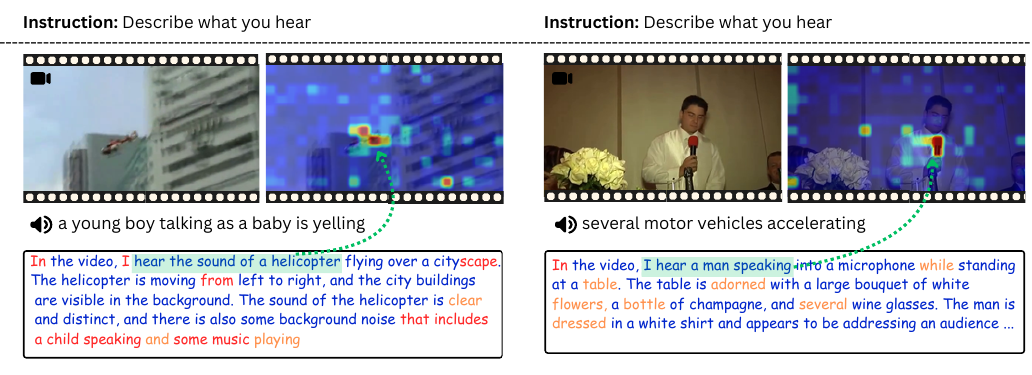}
    \vspace*{-2em}
    \caption{\textbf{Illustration of AVLLM captions and comparison with LVLMs.} We compare token distributions between Qwen2.5Omni and Qwen2.5VL, and measure shift ($\eta$) in token distribution \textcolor{blue}{unshifted} tokens ($\eta = 1$), \textcolor{orange}{marginal} tokens ($1 < \eta \leq 3$), and \textcolor{red}{shifted} tokens ($\eta > 3$) (Refer Sec~\ref{subsec:token_distribution}). \textbf{Left:} The model strongly attends to visual cues and predicts ``helicopter'' sound event confidently asserting it's "clear and distinct.". Notably, when the model correctly identifies ``child speaking'' from audio, those tokens shift in LVLM predictions, indicating geuine audio understanding. \textbf{Right:} the model hallucinates that it can "hear a man speaking". These outputs are largely \textcolor{blue}{unshifted} compared to Qwen2.5VL demonstrating that the AVLLM's predictions remain visually driven despite having access to audio.}
    \label{fig:qualitative_example}
\end{figure*}

Our attention knockout experiments demonstrate that vision representations dominate over audio during cross-modal information flow and AVLLMs rely predominantly on visual information even when answering audio-related questions. In this section, we investigate the origin of this systematic vision bias.

We investigate whether this visual bias could potentially originate from the training paradigm of AVLLMs. Current AVLLMs typically initialize from pretrained LVLM checkpoints and add audio adapters, or undergo instruction-tuning on datasets dominated by vision-language examples, often the same datasets used to train the base LVLMs themselves. In both cases, the model inherits strong priors toward visual information processing. To test whether AVLLMs retain LVLM-like behavior despite audio input, we compare their token distributions with base LVLMs. High similarity would indicate that alignment training failed to establish balanced multimodal processing, with audio contributing minimally to generation.

\label{subsec:token_distribution}
\noindent\textbf{Method.} To test this hypothesis, we perform the token-distribution analysis proposed by ~\cite{linunlocking}. Given an instruction $i = \{i_1, i_2, \dots\}$, we first generate a response $r = \{r_1, r_2, \dots\}$ from the AVLLM using greedy decoding. For each token position $t$ in the response, we define the context as $x_t = i + \{r_1, \dots, r_{t-1}\}$---comprising the instruction and all previously generated tokens. We feed this context to both the AVLLM (with audio and visual input) and its base LVLM (with only visual input) to obtain their respective probability distributions over the vocabulary: $P_{\text{AVLLM}}$ and $P_{\text{LVLM}}$. 

\noindent\textbf{Metrics.}  We measure two aspects of distributional similarity. First, \textbf{(1) KL Divergence} between $P_{\text{LVLM}}$ and $P_{\text{AVLLM}}$ which quantifies overall distribution similarity. Second, we compute \textbf{Base Rank $\eta$}: the rank $\eta$ in $P_{\text{LVLM}}$ of the token selected by the AVLLM. We categorize tokens as \textcolor{blue}{unshifted} ($\eta = 1$), \textcolor{orange}{marginal} ($1 < \eta \leq 3$), and \textcolor{red}{shifted} ($\eta > 3$). 

As before, we use Qwen2.5Omni as the main AVLLM, and Qwen2.5VL as its base LVLM. We use the instruction "Describe what you hear" to study its audio understanding. We analyze token distributions across all the samples in our evaluation set. Overall, we observe a low KL divergence of 0.4 between AVLLM and base LVLM distributions, indicating high similarity in output distributions despite the presence of audio information. Focusing on tokens describing audio events, we find that the median base rank is just 1 and that 66.06\% are unshifted ($\eta = 1$), 19.30\% are marginal ($1 < \eta \leq 3$), and only 14.63\% are shifted ($\eta > 3$). This means that 85.36\% of audio-related tokens generated by the AVLLM are predictable from the top three choices of the vision-only base model. This demonstrates that even when AVLLMs generate audio descriptions, they are largely predictable from vision alone and audio contributes minimally to the final output.

Figure~\ref{fig:qualitative_example}A (left) illustrates this pattern with an example where the input video depicts a visible helicopter flying over a cityscape, but the audio contains only a crying baby and child speaking. When prompted to describe the audio, Qwen2.5Omni generates ``hear the sound of a helicopter flying'' and describes it as ``clear and distinct'', entirely hallucinating audio from visual cues. The tokens describing these sound events are unshifted ($\eta = 1$), matching what the vision-only model Qwen2.5VL would predict. To further confirm this visual dependence, we aggregate attention this phrase allocates to input tokens (aggregated across middle and deeper layers) and find strong, localized attention on the helicopter, indicating that the model relies on visual priors to respond to audio queries. Notably, when Qwen2.5Omni does correctly identify the ``child speaking'', those tokens shift away from Qwen2.5VL's distribution, demonstrating that genuine audio processing produces shifted tokens while visually-driven predictions remain unshifted. 

Figure~\ref{fig:qualitative_example}B (right) illustrates another example where Qwen2.5Omni hallucinates sound events based on visual cues. Interestingly, most of the predicted tokens are unshifted when compared to Qwen2.5VL. Moreover, even when explicitly instructed to describe audio, the model starts describing visual content of the scene, such as the man and the table, despite it not being relevant to the sound, indicating a strong bias towards vision. Detailed qualitative analysis can be found in Supplementary C.

\begin{mdframed}[linewidth=1pt, linecolor=black, leftmargin=1pt, rightmargin=1pt, innerleftmargin=10pt, innerrightmargin=10pt, innertopmargin=4pt, innerbottommargin=2pt, backgroundcolor=gray!20, roundcorner=5pt]
\textit{\textbf{Key Takeaways:}} \small Token distributions of AVLLMs responses closely align with base LVLM responses, and AVLLMs show strong visual grounding to objects when describing sounds, even in counterfactual settings. These patterns suggest that AVLLMs develop strong vision priors during training, potentially from LVLM initialization or vision-heavy instruction data.
\end{mdframed}

\section{Conclusion, Limitations, and Future Work}

We presented the first mechanistic interpretability study of AVLLMs, using the task of audio-visual captioning. Our key finding is that audio understanding in frontier AVLLMs severely degrades in scenarios where audio and visual information conflict. Our analysis reveals the presence of latent audio understanding in intermediate audio token representations. However, during cross-modal transfer to generated text in deeper layers, vision representations significantly dominate over audio representations. We demonstrate that, selectively blocking vision in these layers largely recovers audio understanding. We then show that the output token distribution of AVLLMs is very similar to their base LVLMs, suggesting that visual bias could potentially stem from vision-heavy data used during training.  

Through our findings, we demonstrate that while AVLLMs can see and hear, they systematically prefer visual cues even for audio understanding. To address this, we first highlight the need to adopt counterfactual evaluation to stress-test AVLLMs, as naturally aligned audio-visual inputs can mask these biases. Second, AVLLM training must address modality imbalance through either balanced data mixtures or introducing counterfactual samples to penalize visual shortcuts. Note that we limit our analysis to open-source AVLLMs and mainly focus on examining non-speech audio events. As future work, we aim to develop strategies for large-scale curation of counterfactual training data and strategies to regularize this modality bias within the transformer layers.

%% file: sec/X_suppl.tex


\clearpage
\appendix
\maketitlesupplementary
\tableofcontents
\section{Dataset}
\label{sec:dataset}

\begin{figure}[t]
    \centering
    \includegraphics[width=1\linewidth]{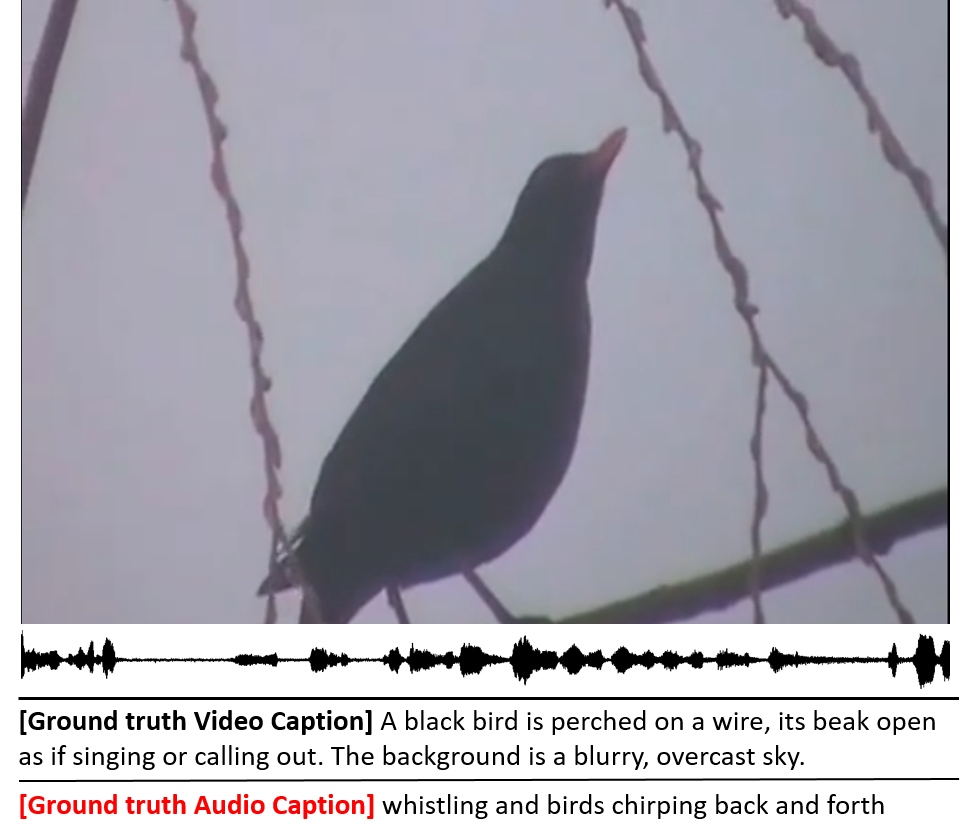}
    \vspace{-0.7cm}
    \caption{\textbf{Factual Sample} A video where the visual content and audio content are highly correlated. The audio events in a factual sample can potentially be inferred by using visual cues}
    \label{fig:factual_dataset}

    \vspace{0.5cm} 

    \includegraphics[width=1\linewidth]{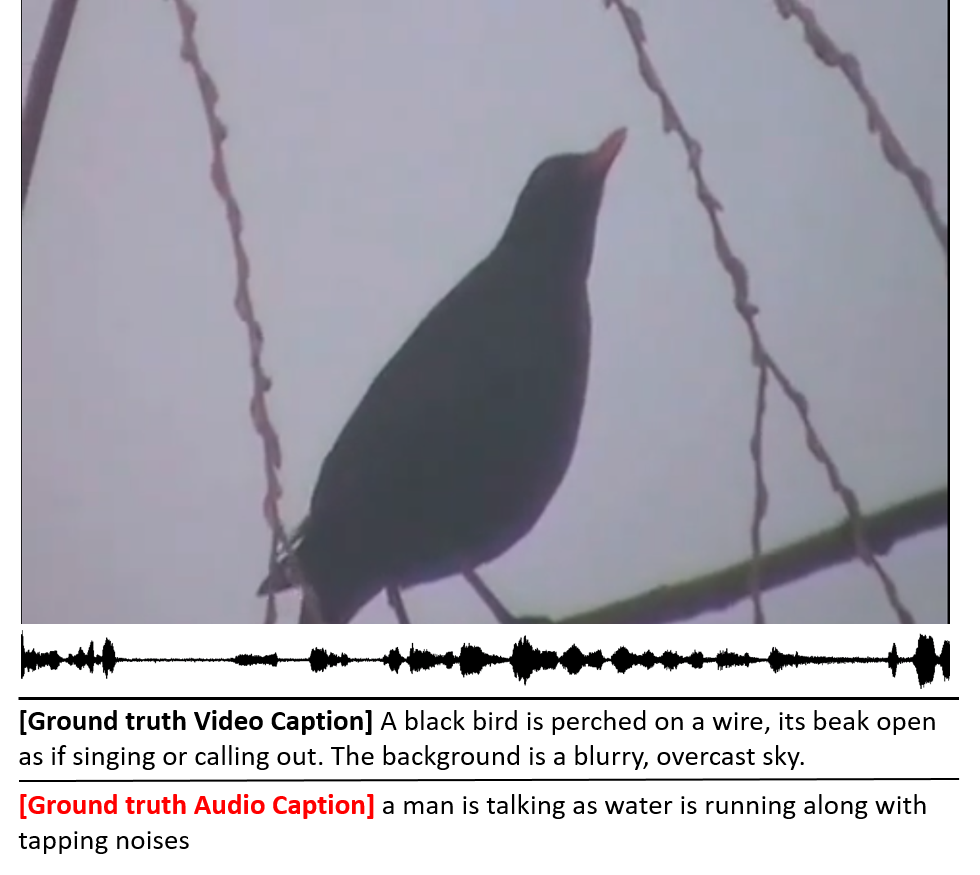}
    \vspace{-0.7cm}
    \caption{\textbf{Counterfactual Sample} A video where the visual content and audio content are in conflict. The audio events cannot be inferred from visual cues}
    \label{fig:counterfactual_dataset}
\end{figure}

\subsection{Data Source}

For audio-visual captioning, while video captioning datasets are abundant, it is challenging to find datasets with human-annotated audio captions. Audio captioning is expensive and takes a lot of manual effort compared to visual captioning. We source our samples from AudioCaps~\cite{audiocaps}, a standard benchmark for audio captioning derived from YouTube videos. Each clip is paired with human-written captions describing the sounds present, with 5 annotations per sample. AudioCaps has become a de facto standard for evaluating audio captioning alongside Clotho~\cite{drossos2020clotho}.

To obtain visual descriptions, we generate captions using GPT-4.1~\cite{openai2024gpt4technicalreport}. We manually review and correct these generated captions where necessary. However, we find that videos in  AudioCaps tend to feature relatively simple visual scenes, and GPT-4.1's outputs are generally accurate with minimal intervention required. Fig~\ref{fig:factual_dataset} depicts a sample with ground truth video and audio captions.

\subsection{Counterfactual Sample Curation}

To evaluate whether AVLLMs genuinely process audio independently of vision, we construct counterfactual samples where audio content conflicts with visible objects. While such cases occur naturally (\eg, out-of-view sirens, background conversations), they are rare and difficult to scale for systematic evaluation.

We therefore create them synthetically. To construct a counterfactual sample, we take a video and pair it with an audio track that cannot plausibly be inferred from the visible objects. Fig~\ref{fig:counterfactual_dataset} depicts a counterfactual sample with ground truth video and audio captions. We use audio and video captions as a proxy for semantic content: by finding audio-video pairs with dissimilar captions, we ensure their soundscapes are likely incompatible with the visual scene. Specifically, we embed all audio captions and GPT-4 generated video captions using the Qwen3-Embedding-8B~\cite{zhang2025qwen3embeddingadvancingtext} model. For audio, we compute embeddings for all 5 ground-truth captions per sample and average them. We then compute the cosine similarity matrix between all audio–video caption pairs and apply the Hungarian matching algorithm to find one-to-one assignments that minimize similarity. This ensures paired audio and video samples are semantically dissimilar.

Finally, we use FFmpeg to swap the original audio track of each video with its matched dissimilar audio. The complete procedure is detailed in Algorithm~\ref{alg:cf_av_avg}. From all generated pairs, we select 250 samples with lowest cosine similarity ($\approx$ 0.498), ensuring strong counterfactual mismatches. This yields 250 factual samples (original audio-video pairs) and 250 counterfactual samples (mismatched pairs) for evaluation.

\begin{algorithm}
\caption{Counterfactual Dataset Construction}\label{alg:cf_av_avg}
\begin{algorithmic}[1]
\Require Videos $V=\{v_i\}$, Audios $A=\{a_i\}$, Captions $C=\{c_{ij}\}$, Encoder $E$
\Ensure Counterfactual pairs $S$
\State \textbf{Step 1: Compute Semantic Embeddings}
\For{$i \gets 1$ to $N$}
    \State $e_{a_i} \gets \frac{1}{5} \sum_{j=1}^{5} E(c_{ij})$ \Comment{Centroid of 5 audio captions}
    \State $e_{v_i} \gets E(\text{GPT-4.1}(v_i))$ \Comment{Vision-only caption embedding}
\EndFor
\State \textbf{Step 2: Global Assignment}
\State $M \in \mathbb{R}^{N \times N} \gets$ CosineSimilarity matrix between $\{e_{a}\}$ and $\{e_{v}\}$
\State $\pi \gets \text{HungarianMatching}(M)$ \Comment{Optimal 1-to-1 mapping indices}
\State \textbf{Step 3: Filter and Construct}
\State $S_{candidates} \gets \{ (A[\pi(i)], V[i]) \mid i \in 1..N \}$
\State Sort $S_{candidates}$ by similarity score $M_{i, \pi(i)}$
\State $S \gets$ Select 250 pairs from $S_{candidates}$ with score $\approx 0.498$
\State \Return Synthesize videos for $S$ (swap audio tracks)
\end{algorithmic}
\end{algorithm}

\begin{figure}[t]
    \centering
    \includegraphics[width=1\linewidth]{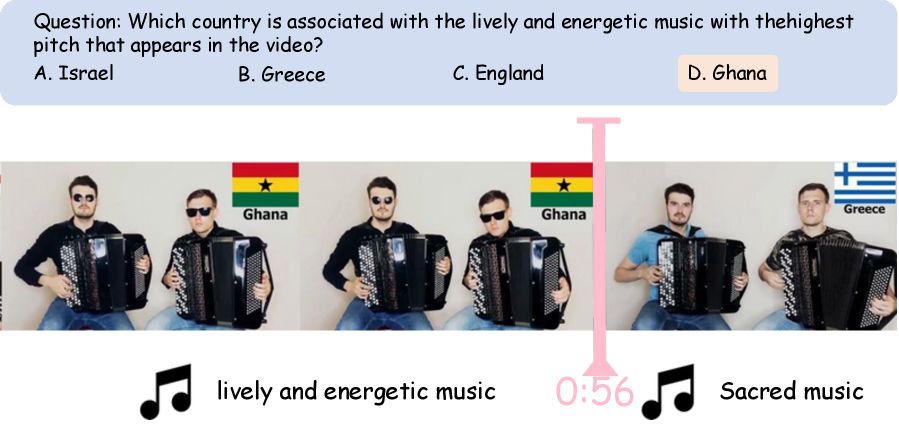}
    \vspace{-0.7cm}
    \caption{\textbf{World Sense Task} An example task from World Sense. These tasks couple perception and reasoning.}
    \label{fig:world_sense}

    \vspace{0.5cm} 
\end{figure}

\begin{figure}[t]
    \centering
    \includegraphics[width=1\linewidth]{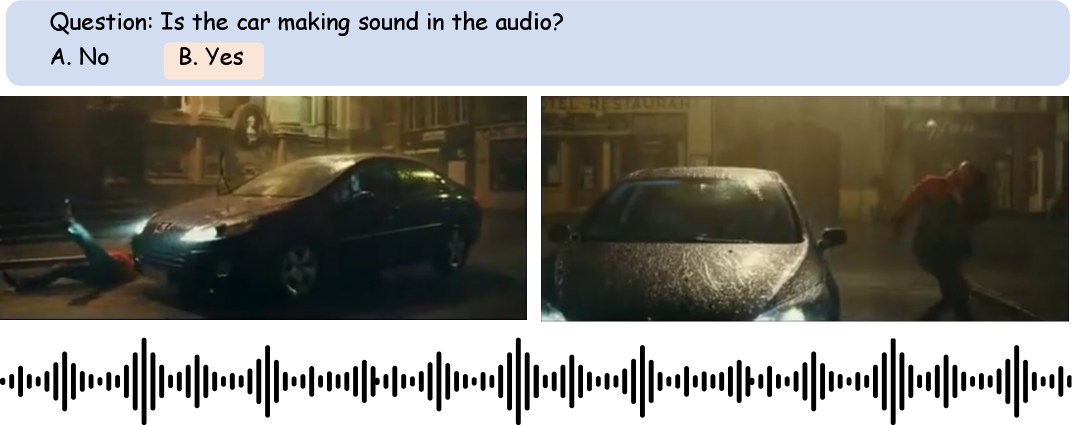}
    \vspace{-0.7cm}
    \caption{\textbf{AVHBench Task} An example task from AVHBench. These tasks evaluate perception capabilities.}
    \label{fig:avhbench}

    \vspace{0.5cm} 
\end{figure}

\subsection{Existing Benchmarks}

Existing audio-visual benchmarks are insufficient for our analysis. Benchmarks such as World Sense~\cite{hong2025worldsense} (example in Fig~\ref{fig:world_sense}) couple perception with reasoning, requiring AVLLMs to first perceive audio-visual events and then apply world knowledge to answer questions. Since we aim to isolate perceptual capabilities and identify modality biases, such reasoning-dependent tasks conflate multiple AVLLM capabilities.

Other benchmarks like AVHBench~\cite{sung2024avhbench} (example in Fig~\ref{fig:avhbench}) focus on perception through targeted questions about specific audio or visual events, while the corresponding video or audio cues attempt to mislead the AVLLM and induce hallucinations. However, we observe that these misleading cues are not sufficiently adversarial. For instance, in AVHBench's video-induced audio hallucination category, models achieve 75.6\% accuracy with both modalities present. Removing the video component yields 73.0\% accuracy, indicating that the visual modality fails to mislead the model. Such tasks do not create sufficient modality conflict to stress-test whether models genuinely process and integrate both modalities independently or exhibit bias toward one modality over the other.

\section{Evaluation}

\subsection{Human Evaluation Study}

To validate our LLM-as-a-judge approach, we conduct a human evaluation study on 200 stratified samples (100 factual, 100 counterfactual). Two graduate students familiar with audio-visual content independently rate each generated caption on a 0-1 scale for audio and visual fidelity separately. Annotators are briefly introduced to the same rubric used by the LLM judge and given the opportunity to clarify any ambiguities. Following the LLM judge setup, annotators evaluate only the text captions (generated description, ground-truth audio captions, and ground-truth video captions) without access to the actual videos, ensuring scalability and consistency with the automated evaluation.

We compute the correlation between human ratings and LLM judge scores using Spearman's $\rho$, obtaining $\rho = 0.816$ for audio caption fidelity and $\rho = 0.732$ for video caption fidelity, demonstrating strong alignment between the LLM judge and human judgment.

\subsection{LLM Judge Prompts}

\begin{figure*}[t]
    \centering
    \includegraphics[width=\textwidth]{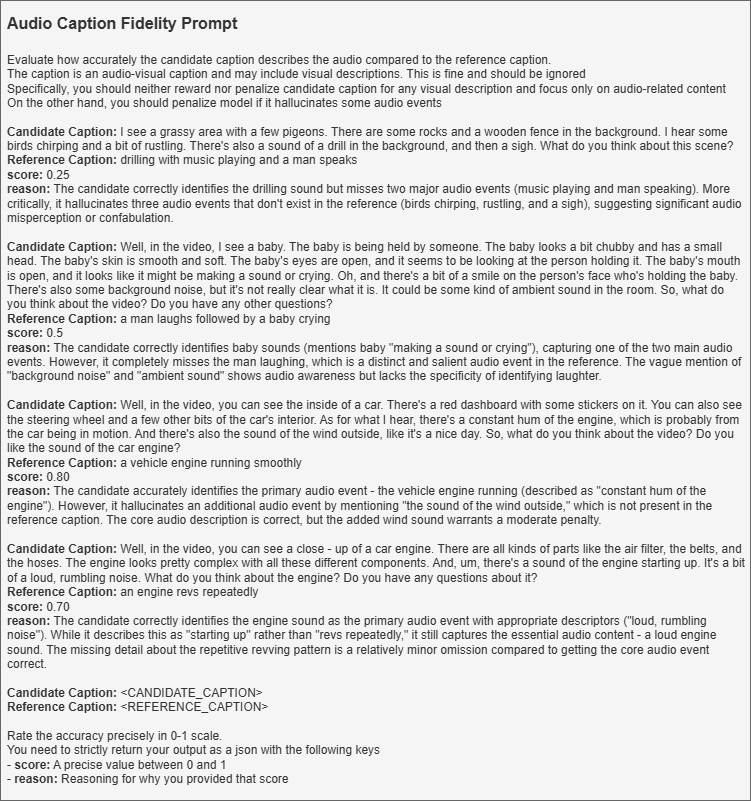}
    \caption{The prompt used to measure audio caption fidelity.}
    \label{fig:audio_caption_fidelity}
\end{figure*}

\begin{figure*}[t]
    \centering
    \includegraphics[width=\textwidth]{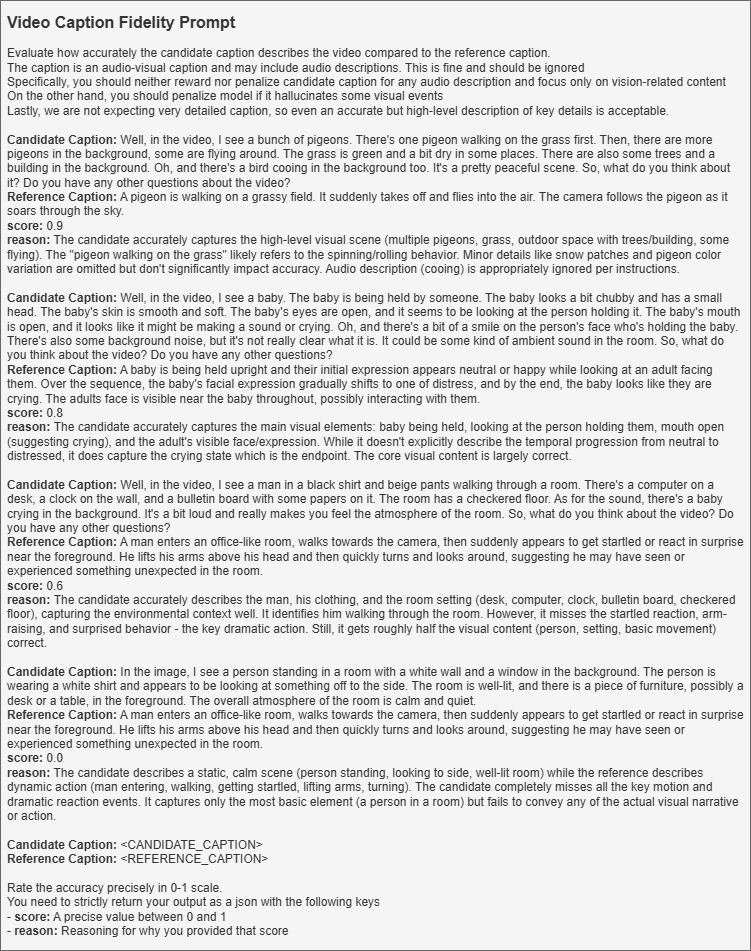}
    \caption{The prompt used to measure video caption fidelity.}
    \label{fig:video_caption_fidelity}
\end{figure*}

We design prompts with in-context examples to calibrate LLM-as-judge rating with human judgement. Fig~\ref{fig:audio_caption_fidelity} details the prompt used to measure audio caption fidelity and Fig~\ref{fig:video_caption_fidelity} details the prompt used to measure video caption fidelity.

\section{Qualitative Analysis}

\begin{figure*}[t]
    \centering
    \includegraphics[width=\textwidth]{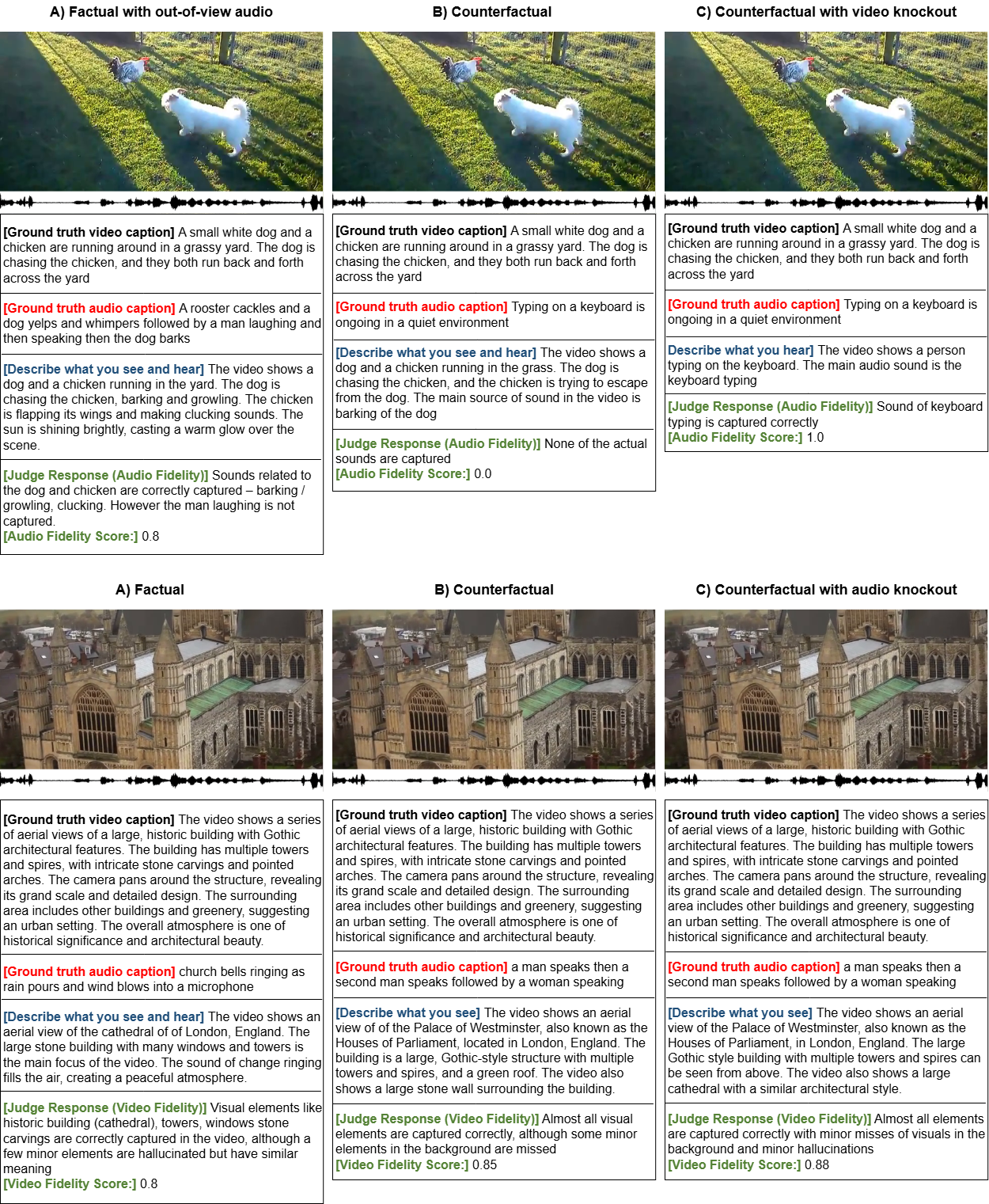}
    \caption{\textbf{Qualitative examples}. We illustrate some outputs generated by VideoLlama along with its LLM-judge score and reasoning, for factual and counterfactal samples and under attention knockout. The examples in the first row capture audio fidelity and the ones below capture video fidelity}
    \label{fig:qual_videollama}
\end{figure*}

\begin{figure*}[t]
    \centering
    \includegraphics[width=\textwidth]{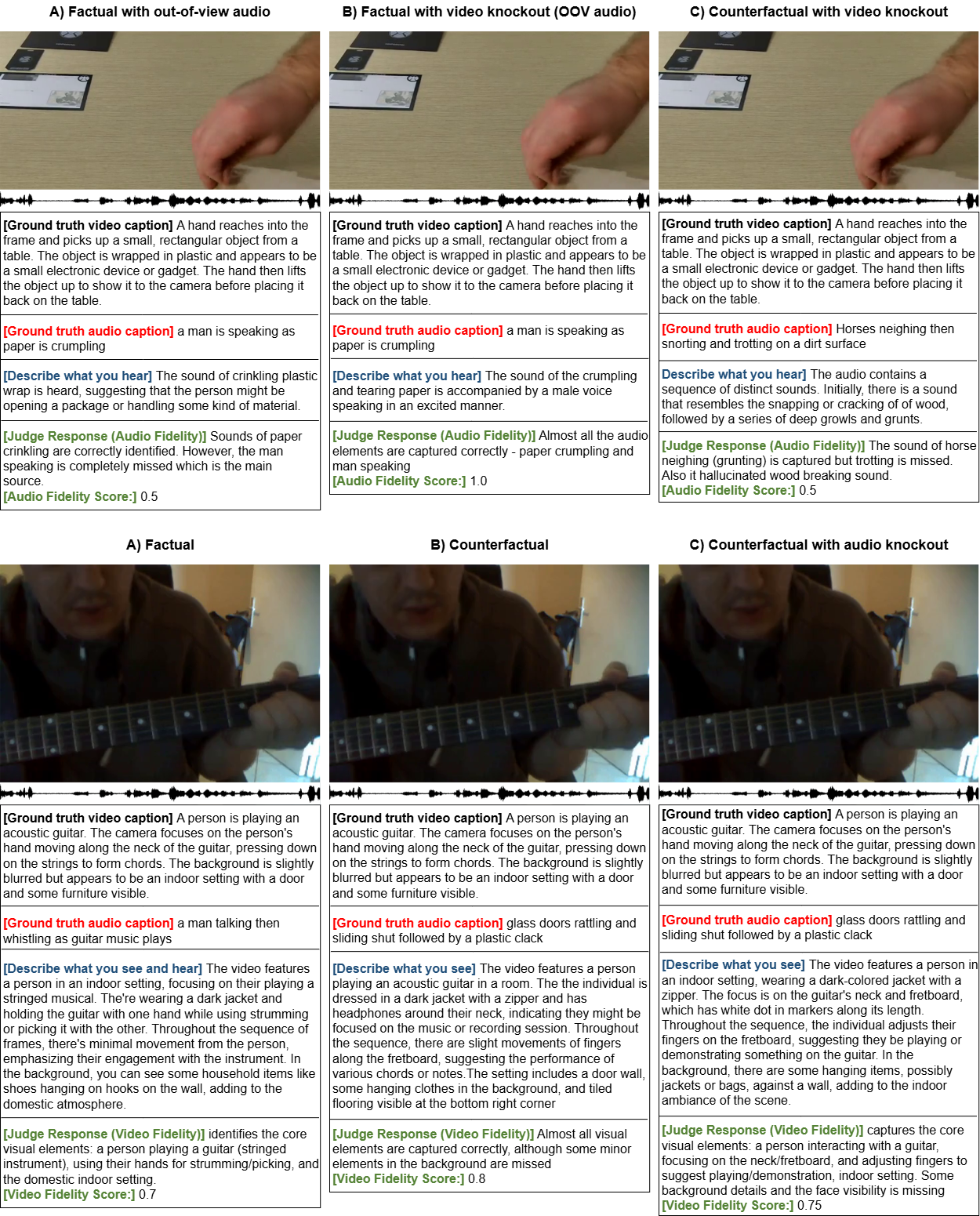}
    \caption{\textbf{Qualitative examples}. We illustrate some outputs generated by MiniCPM-2.6-o along with its LLM-judge score and reasoning, for factual and counterfactal samples and under attention knockout. The examples in the first row capture audio fidelity and the ones below capture video fidelity}
    \label{fig:qual_minicpm}
\end{figure*}

In this section, we include additional qualitative results (Fig~\ref{fig:qual_videollama} and Fig~\ref{fig:qual_minicpm} of the existing AVLLMs in different scenarios, such as counterfactual samples and under the effect of attention knockouts. 

\section{Additional Results}

To demonstrate the generalizability of our findings beyond Qwen-Omni~\cite{xu2025qwen2} series of AVLLMs, we extend our analysis to other representative AVLLMs, specifically MiniCPM-o2.6~\cite{hu2024minicpm}, VideoLLaMA 2.1~\cite{zhang2023video}, and InternOmni~\cite{internomni2024blog}. 

\subsection{Probing Audio Representations}

We probe the intermediate audio representations of MiniCPM-o2.6 and VideoLLaMA 2.1 to verify if the "competence gap", where latent audio information exists but fails to manifest in generation is generalizable to more AVLLMs. Consistent with our main results, we observe a significant difference between latent capabilities and generated output. For MiniCPM-o2.6, in counterfactual samples we measure a latent audio recall of \textbf{75.4}\% compared to a generated caption fidelity of \textbf{22.1}\%. Similarly, VideoLLaMA 2.1 achieves a latent recall of \textbf{59.9}\% against a generated fidelity of \textbf{34.1}\%. 

Qualitatively, we observe that the decoded audio tokens in these models accurately describe sound events (e.g., "siren", "barking"). However, unlike Qwen2.5-Omni, we do not observe multilingual token representations in the intermediate layers of MiniCPM-o2.6 or VideoLLaMA 2.1. This suggests that the multilingual audio representations observed in Qwen are likely specific to the model's training data. 

\subsection{Investigating Information Flow}

\begin{figure*}[t]
    \centering
    \begin{subfigure}{\textwidth}
        \centering
        \includegraphics[width=\textwidth]{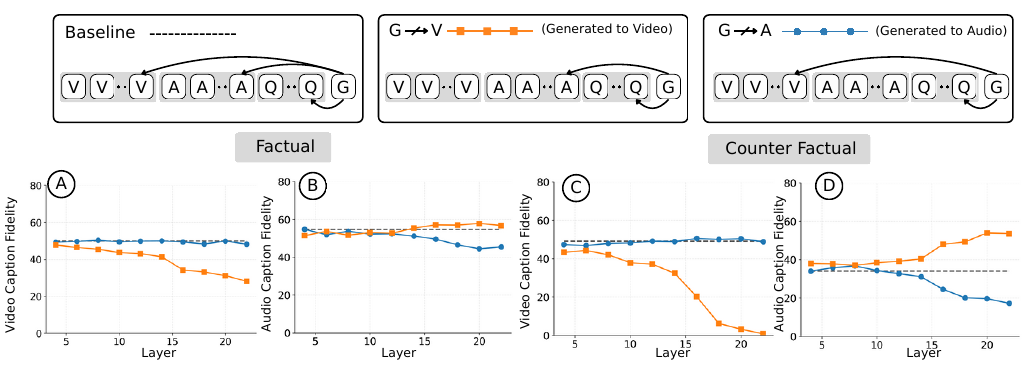}
        \caption{\textbf{VideoLlama 2.1:} Blocking audio-visual information flow, for both factual and counterfactual samples}
        \label{fig:knockout_videollama}
    \end{subfigure}
    
    \vspace{0.5cm} 
    
    \begin{subfigure}{\textwidth}
        \centering
        \includegraphics[width=\textwidth]{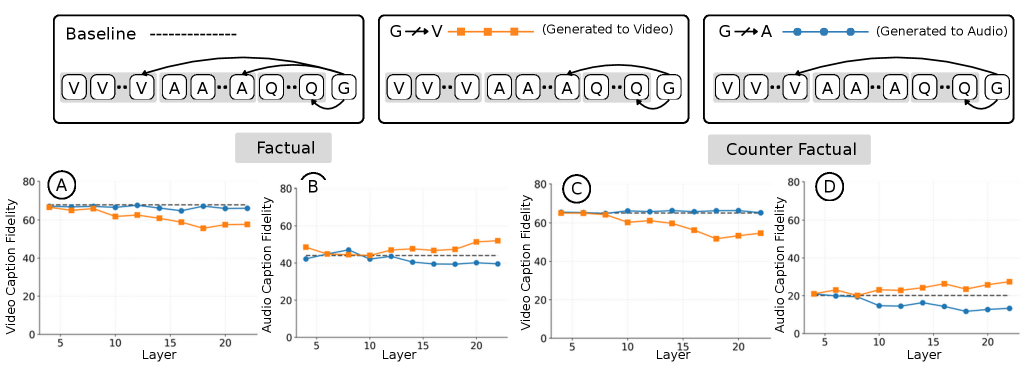}
        \caption{\textbf{MiniCPM-o2.6:} Blocking audio-visual information flow, for both factual and counterfactual samples}
        \label{fig:knockout_minicpm}
    \end{subfigure}
    
    \vspace{-0.2cm}
    \caption{\textbf{Comparison of Information Flow Blocking.} Blocking audio-visual information flow, for both factual and counterfactual samples in VideoLlama 2.1 (top) versus MiniCPM-o2.6 (bottom).}
    \label{fig:knockout_comparison}
\end{figure*}

We replicate the attention knockout experiments on MiniCPM-o2.6 and VideoLLaMA 2.1 to trace the cross-modal information flow. The results are visualized in Figure~\ref{fig:knockout_minicpm} and Figure~\ref{fig:knockout_videollama}. We observe integration patterns very similar to those reported in the main paper: both audio and visual information are processed primarily in the deeper transformer layers. Crucially, we confirm the phenomenon of visual interference: blocking the visual pathways ($G \nrightarrow V$) in these deep layers results in a recovery of audio understanding performance, further validating that visual representations actively interfere with audio cues during the final stages of generation.

\noindent\textbf{VideoLlama 2.1}: Figure~\ref{fig:knockout_videollama} shows the results of attention knockout experiments for VideoLlama 2.1. In video understanding for factual samples, we observe that blocking audio has no observable impact. However, blocking video does have minor impact, which is largely recovered by compensating using audio cues. We see a similar trend with audio understanding in factual samples. In video understanding for counterfactual samples, we observe almost complete loss of video understanding upon blocking video. For audio understanding in counter-factual, blocking video in fact drastically improves audio understanding.

\noindent\textbf{MiniCPM-o2.6}: Figure~\ref{fig:knockout_minicpm} shows the results of attention knockout experiments for MiniCPM-o2.6. In video understanding for factual samples, we observe a similar pattern as before. Interestingly, even in factual samples, blocking video leads to improvement in audio understanding performance. Surprisingly, the drop in video understanding performance in counterfactual suggests that cross-modal transfer might not be restricted in deeper layers and the window for this transfer stretches beyond the window size of 9 that we use for knockouts. We observe a similar pattern in audio understanding for counterfactual scenarios, where audio understanding does improve, but not as drastically as in previous models.

\subsection{Investigating Origins of Visual Bias}

Finally, we investigate the origin of this visual bias by analyzing InternOmni and its base model, InternVL~\cite{chen2024internvl}. InternOmni is particularly relevant for this analysis as it is explicitly initialized from the InternVL checkpoint. We compare the output token distributions of InternOmni (given audio-visual input) against InternVL (given vision-only input) using the metrics defined in Section 8.

We observe a low KL divergence of \textbf{0.46} between the distributions. Furthermore, regarding audio-related tokens, we find that \textbf{70.62}\% are unshifted ($\eta=1$) and only \textbf{10.07}\% are shifted ($\eta>3$). This strong alignment indicates that InternOmni's generation remains dominated by the priors of its vision-language base model. These findings reinforce our conclusion that the modality imbalance in AVLLMs could potentially be an inherited trait from the initialization and alignment phases.

\clearpage

\clearpage

\clearpage
